\documentclass{svmult}
\usepackage[utf8]{inputenc}
\usepackage{amsmath}
\usepackage{algorithm}
\usepackage[]{algpseudocode}
\usepackage{setspace}
\usepackage{parskip}
\usepackage{sectsty}
\usepackage{fancyhdr}
\usepackage{tabularx,ragged2e,booktabs,caption}

\sectionfont{\fontsize{16}{15}\selectfont}
\author{Ivan S. Grechikhin}
\title*{Heuristic with elements of tabu search for Truck and Trailer Routing Problem}

\algnewcommand{\LineComment}[1]{\State \(\triangleright\) #1}
\algnewcommand\textfunc{\textsc}

\begin{document}

\maketitle\footnotetext{National Research University Higher School of Economics. Laboratory of Algorithms and Technologies for Network Analysis. Probationer; E-mail: igrechikhin@hse.ru}

\abstract{Vehicle Routing Problem is a well-known problem in logistics and transportation, and the variety of such problems is explained by the fact that it occurs in many real-life situations. It is an NP-hard combinatorial optimization problem and finding an exact optimal solution is practically impossible. In this work, Site-Dependent Truck and Trailer Routing Problem with hard and soft Time Windows and Split Deliveries is considered (SDTTRPTWSD). In this article, we develop a heuristic with the elements of Tabu Search for solving SDTTRPTWSD. The heuristic uses the concept of neighborhoods and visits infeasible solutions during the search. A greedy heuristic is applied to construct an initial solution. \\}


\section{Introduction}

\setcounter{page}{3}
Vehicle Routing Problem is a well-known problem in combinatorial optimisation and integer programming. The problem can be secribed as follows: there is a set of customers, where each customer has a demand, there is a set of vehicles, which may serve the demand of customers. Using the information on the distances and costs of travelling between each pair of customers, the goal is to find the solution with minimal total cost. This paper considers one version of the problem, which is called Truck and Trailer Routing Problem (TTRP). The problem in consideration is a real-life problem, and contains a big number of constraints. 

Truck and trailer routing problem has two sets of customers: truck-customers and trailer-customers. Every vehicle, then, consists of a truck and a trailer of some capacities (sometimes trailer capacity is zero, which means the vehicle does not have a trailer). Truck-customers can not be served by a vehicle with a trailer. It means that the vehicle should not have trailer from the start, or the trailer should be left at some other place before visiting a truck-customer. This requirement is explained by the fact that there may be small stores, that do not have place for vehicle with its trailer. A vehicle with a trailer has a possibility to leave the trailer at a transshipment location, which is basically a special place to leave trailers. Another opportunity is to leave trailer at a previous trailer-customer: in this case the trailer may be unloaded at the trailer-customer and, at the same time, the truck goes to a truck-customer and serves it in parallel with the trailer-customer. Such rules create a necessity to organize load transfer - the operation, where goods are transferred from truck to trailer or vice versa. This may happen, because, for example, the total weight of goods for truck-customers in one route is more than the capacity of the truck.

In the considered problem, the Heterogeneous Fleet of vehicles is present (HFTTRP). This problem differs from homogeneous fleet TTRP, where all vehicles are the same: they have the same fixed costs and capacities. HFTTRP has a set of vehicles with different capacities and fixed costs, which makes the problem even more difficult. Additionally, every customer may have its own preferences on types of vehicles to serve the customer. In this case the problem is called the Site Dependent TTRP (SDTTRP) and there are some developed heuristics for solving such problems sometimes with additional elaborations in constraints . 

Another real-life constraints are hard and soft time windows and split-deliveries. Time windows are periods of time, when the delivery is acceptable (hard time windows) and the constraint should be satisfied in the majority of routes (soft time window). Split-deliveries are such deliveries, when there is a possibility to serve one customer with more than one vehicle. The problem in this form is considered in Batsyn \& Ponomarenko (2014) and Batsyn \& Ponomarenko (2015). These papers suggested greedy heuristic for the problem. In this article, this heuristic is developed in another way with an addition of new heuristic with elements of tabu-search. The greedy heuristic is altered so there are possible operations of both insertion and deletion from the route. After the greedy heuristic, the obtained solution is reconstructed with new heuristic with tabu-search elements.

Greedy heuristic constructs the solution iteratively, until there are no unserved customers. For every route, the algorithm randomly chooses one of the farthest customers to be the first customer added to the route. Then, other customers are tried as candidates to the route. The solution has a constraint on the number of split-deliveries and delays (violations of soft time window). For every route, the possibility of a split-delivery and the number of delays is chosen randomly. The route may have only one new split-delivery, and the probability is determined by the fraction of the current allowed split-deliveries to the estimated number of split-deliveries. The number of delays is defined by the fraction of the current allowed soft window violations to the estimated number of soft window violations, however, every route may have different number of delays. The algorithm determines the allowed number of delays before constructing the route. After the solution is obtained, the heuristic with tabu-search elements tries to move customers between the routes to derive better solution. The algorithm uses set of changing parameters, which define “tabu neighbourhood” - the algorithm looks through the infeasible solutions. The degree of infeasibility is determined by the parameters - the number of allowed delays over limit, the number of routes with violated capacity and allowed cost change of the route. 

\section{Heuristic algorithm}

The following parameters are used in the pseudo-code of the algorithm:\\
$n$ - the number of customers \\
$V$ - the set of all customers \\
$K$ - the set of all vehicles \\
$K_i$ - the set of vehicles, which can serve customer $i$ \\
$Q_k$ - the current remaining capacity of vehicle $k$ \\
$q_i$ - the current remaining demand of the customer $i$ \\
$v_R$ - the number of soft time window violations in route R \\
$R$ - the current route \\
$S$ - the current solution \\
$S^*$ - the best solution found so far \\
$v$ - the number of permitted soft window violations \\
$w$ - the current remaining number of permitted soft time window  violations \\
$U$ - the set of all customers sorted the most expensive (farthest) customer first \\
$C$ - the cost of current insertion \\
$corridor$ - the allowed level of violations \\
$CurrentState$ - the current state of second heuristic \\
$closeness$ - the distance between customers to perform move \\
$CVset$ - the set of routes, where the capacity of vehicles is violated \\

\begin{algorithm}
\caption{Initial Greedy heuristic Part 1}\label{euclid}
\begin{algorithmic}[1]
\Function{InitialGreedyHeuristic}{}

\LineComment Creates one initial feasible solution
\State $U \gets V$ \Comment sorting customers so that $U_1$ has maximal
$c_{0i}^{kl}$
\State $S \gets \emptyset$
\While{$U \neq \emptyset$}
\State $w = v$ 
\State $i \gets $\textfunc{Random}$(U_1,\dots,U_\mu)$ \Comment choose from the $\mu$ most expensive
\State $k \gets $\Call{ChooseVehicle}{$i, \left[q_j\right], \left[Q_k\right]$}
\State $R \gets $\Call{BasicRoute}{$k$}
\State $violNumber \gets $\Call{FindNumberOfViolations}{$i,k,w$}
\State $R_{viol} \gets \emptyset,\hspace{5pt}R_{clear} \gets \emptyset$
\State $C_{viol} \gets \infty\hspace{5pt}C_{clear} \gets \infty$
\State $ID_{viol} \gets 0, \hspace{5pt}ID_{clear} \gets 0$
\For{$j \in U$}
\If{$k \notin K_j$}
\State \textbf{continue}
\EndIf
\State $mayViolate \gets true$
\State $C_{viol}^{'} \gets $\Call{GetInsertionCost}{j,R,mayViolate,$q_j$,$Q_k$,$violNumber$}
\State $mayViolate \gets false$
\State $C_{clear}^{'} \gets $\Call{GetInsertionCost}{j,R,mayViolate,$q_j$,$Q_k$,$violNumber$}
\LineComment There are two possible insertions, with violation or without
\If{$C_{viol}^{'} < C_{viol}$}
\State $C_{viol} \gets C_{viol}^{'}$
\State $R_{viol} \gets $\Call{InsertCustomer}{j,R,$true$,$q_i$,$Q_k$}
\State $ID_{viol} \gets j$
\EndIf
\If{$C_{clear}^{'} < C_{clear}$}
\State $C_{clear} \gets C_{clear}^{'}$
\State $R_{clear} \gets $\Call{InsertCustomer}{j,R,$false$,$q_i$,$Q_k$}
\State $ID_{clear} \gets j$
\EndIf

\algstore{break}

\end{algorithmic}

\end{algorithm}

\begin{algorithm}[h]
\caption{Initial Greedy heuristic Part 2}\label{euclid}
\begin{algorithmic}[1]
\algrestore{break}
\If{$R_{viol} = null \hspace{10pt} and \hspace{10pt} R_{clear} = null$}
\State $S \gets S \cup \left\{R\right\}$
\State $Q_k \gets 0$
\State $w \gets w - v_R$ 
\State \textbf{break}
\ElsIf{$R_{clear} = null$}
\State $R \gets R_{viol}$
\State $U \gets U\slash \{ID_{viol}\}$
\ElsIf{$R_{viol} = null$}
\State $R \gets R_{clear}$
\State $U \gets U\slash \{ID_{clear}\}$

\Else
\If{$C_{viol} > C_{clear}$}
\State $R = R_{viol}$
\State $U \gets U\slash \{ID_{viol}\}$
\Else
\State $R = R_{clear}$
\State $U \gets U\slash \{ID_{clear}\}$
\EndIf
\EndIf
\EndFor
\EndWhile

\EndFunction
\end{algorithmic}
\end{algorithm}

\begin{algorithm}
\caption{Heuristic with Tabu Search}\label{euclid}
\begin{algorithmic}[1]
\Function{TabuSearchHeuristic}{$S$,$corridor$,$closeness$,$CurrentState$}
\State $S^* \gets S$
\Repeat
\State $success \gets $\Call{HeuristicStep}{$S$,$corridor$,$closeness$}
\State \Call{ChangeStateForTabuStepSuccess}{$CurrentState$,$success$}
\If{\Call{ShouldObtainFeasibleSolution}{$CurrentState$}}
\LineComment Recovery procedures work here
\State \Call{RoutesOptimization}{$S$}
\State \Call{RecoverCapacityViolations}{$S$}
\State \Call{FinalzeRoutesTimes}{$S$}
\State \Call{RecoverSoftWindowViolations}{$S$}
\If{\Call{Cost}{$S$} $<$ \Call{Cost}{$S^*$}}
\State $S^* \gets S$
\State \Call{ChangesStateForChangeInBest}{$CurrentState$}
\EndIf
\EndIf
\State \Call{ChangeCorridor}{$corridor$,$CurrentState$}
\Until{\Call{StoppingCondition}{$CurrentState$}}
\EndFunction
\end{algorithmic}
\end{algorithm}

The first important function of the whole algorithm is initial greedy heuristic, which constructs initial solution (Algorithms 1 and 2). The function works so that the solution will be necessarily constructed, but its cost may not be satisfactory. First, the algorithm sorts all customers by the distance from the depot (or, by the cost of direct travel from depot, which is the same) so the first customer in $U$ is the farthest. Then, the process of solution construction begins. Routes of the solution are constructed in cycle, until there are unserved customers. For every route, the algorithm chooses one of the farthest customers, after that the vehicle is determined for the route. Also, function \Call{BasicRoute}{$k$} creates the route with one chosen customer. 

The function \Call{FindNumberOfViolations}{$i,k,w$} determines maximal possible number of soft window violations for the current route. The function uses the relation of current remaining soft window violations to the estimated number of remaining soft window violations and increases the number of allowed violations until the random generator returns numbers less than this relation. After that, the algorithm tries to insert all other customers in the route $R$, however, the algorithm does the insertion in two ways - allowing the violation of soft time window and forbidding the violation. If the number of soft window violations exceeds the allowed number, the route is forbidden. From obtained routes, there is chosen the best. Step by step the algorithm inserts customers until there are no possible insertions. 

The whole idea of the greedy algorithm is based on Batsyn \& Ponomarenko (2014) and Batsyn \& Ponomarenko (2015)

Second important function is the second heuristic with elements of tabu search (Algorithm 3). Its goal is to take initial solution $S$ and improve it by performing simple moves. The algorithm makes steps and at each step there is a possible move happens. The variety of possible moves depends on the $corridor$ and $closeness$ parameters. Also, there is $CurrentState$ of the algorithm, which tracks successes, changes in the current best and some other parameters. From time to time, the algorithm tries to obtain feasible solution from current solution. The algorithm also may change $corridor$ depending on $CurrentState$ of the heuristic or even stop it in order to get new initial solution and start the procedure again.

At every step of the second heuristic (Algorithm 4), first, the customer is chosen randomly from one of the route of the current solution. After that, the customer is tried to be inserted in other routes in such way that the adjacent customer is close - the time of travel is less than $closeness$ parameter. Variables $places$ contains all such places of insertion in the route $R$. After finding the best move by the cost this move may be performed if it does not violate too many constraints. 

\begin{algorithm}
\caption{Heuristic Step Algorithm}\label{euclid}
\begin{algorithmic}[1]
\Function{HeuristicStep}{$S$,$corridor$,$closeness$}
\State $R_i,i,costOfDeletion \gets $\Call{ChooseRandomCustomer}{$S$}
\LineComment $i$ is deleted customer, the algorithm also needs the cost of deletion of this customer from its current route
\State $bestCost \gets \infty$
\State $bestRoute \gets \emptyset$
\ForAll{$R \in S$}
\State $places \gets$\Call{FindPlacesForInsertion}{$R$,$closeness$}
\ForAll{$place \in places$}
\State $R^* \gets $\Call{AddCustomer}{$R$,$place$}
\State $cost \gets $\Call{FindMoveCost}{$costOfDeletion$,$R$,$R^*$,$corridor$,$CurrentState$}
\If{$cost < bestCost$}
\State $bestCost = cost$
\State $bestRoute = R$
\EndIf
\EndFor
\EndFor
\State $success \gets $\Call{AllowMove}{$bestCost$,$corridor$}
\If{$success$}
\State $R_i \gets R_i / {i}$
\State $R \gets R \cup {i}$
\State \Call{ChangeCurrentViolations}{$CurrentState$,$S$}
\EndIf
\EndFunction
\end{algorithmic}
\end{algorithm}

\begin{algorithm}
\caption{Recovery Capacity Violations Procedure Part 1}\label{euclid}
\begin{algorithmic}[1]
\Function{RecoverCapacityViolations}{$S$}
\State $CVset \gets $\Call{FindRoutesWithCapacityViolations}{$S$}

\ForAll{$R \in CVset$}
\State $bestCost \gets \infty$
\State $bestCustomer \gets -1$
\State $bestRouteFrom \gets \emptyset$
\State $bestRouteTo \gets \emptyset$
\ForAll{$i \in R$}
\ForAll{$R^c \notin CVset$}
\State $cost,R^*, R^{c*} = $\Call{FindCostOfMove}{$R$,$i$,$R^c$}
\If{$cost < bestCost$}
\State $bestCost \gets cost$
\State $bestCustomer \gets i$
\State $bestRouteFrom \gets R^*$
\State $bestRouteTo \gets R^{c*}$
\EndIf
\EndFor
\EndFor




\If{$bestCost \neq \infty$}
\State \Call{ReplaceRoutes}{$S$,$bestRouteFrom$,$bestRouteTo$}
\EndIf
\EndFor
\State $CVset = $\Call{FindRoutesWithCapacityViolations}{$S$}
\While{$CVset \neq \emptyset$}
\State $R_{cap} \gets $\Call{ChooseRandomRoute}{$CVset$}
\State $i \gets $\Call{ChooseCustomerToRecoverCapacity}{$R_{cap}$}
\State $CVset \gets CVset \slash R_{cap}$
\State $success \gets (CVset \neq \emptyset)$
\While{$R \in CVset$}
\State $bestCost \gets \infty$
\State $bestCustomer \gets -1$
\State $bestRouteFrom \gets \emptyset$
\State $bestRouteTo \gets \emptyset$
\ForAll{$i \in R$}
\ForAll{$R^c \notin CVset$}
\State $cost,R^*, R^{c*} = $\Call{FindCostOfMove}{$R$,$i$,$R^c$}
\If{$cost < bestCost$}
\State $bestCost \gets cost$
\State $bestCustomer \gets i$
\State $bestRouteFrom \gets R^*$
\State $bestRouteTo \gets R^{c*}$
\EndIf
\EndFor
\EndFor
\EndWhile
\EndWhile
\EndFunction
\end{algorithmic}
\end{algorithm}

Finally, when the second heuristic tries to obtain the feasible solution from current infeasible, the recovery procedure takes place(Algorithm 3). Basically, the whole solution is likely to be in infeasible region because of moves. In that case, the algorithm needs to decrease the number of soft time window violations and recover over-capacitated routes to be under constraints. The recovery procedures start with route optimization - it creates some free space inside routes in order to recover solution more efficiently. After that, the algorithm recovers capacities of routes. Next step is finalization of times - the procedure goes through every route and compacts the time of the route. The last step is recovering soft time window violations. 

The algorithm of capacity constraints recovery is described in Algorithm 5. There are two parts in this algorithm. First part of the algorithm repeatedly tries to take customers from over-capacitated routes and insert them in other routes without capacity violations. If there is no such move possible and there are over-capacitated routes left, the second part of the algorithm creates new routes with customers from over-capacitated routes. At the end of the procedure all routes have total demand less or equal to the capacity of the vehicle of the route.

\section{Computational results}

Experiments were performed for seven experimental days, for which the good results of greedy heuristic are known. The column Greedy Heuristic Results, contains the value of objective function obtained by the greedy heuristic for this day (Batsyn \& Ponomarenko, 2015). The third column shows the results of heuristic with tabu search elements for the experimental days. The second heuristic worked for 3 hours for every experimental day. All experiments were conducted on Intel Xeon X5675 machine, with base processor frequency 3.06 GHz and 64 GB of memory.

\begin{center}
\captionof{table}{Computational results}
\begin{tabular}{| c | c | c | c |}
\hline
Day & Greedy Heuristic Results & Tabu Search Heuristic Results & Improvement \\ \hline
Day 1 & 1200000 & 1155000 & -4\% \\ \hline
Day 2 & 1100000 & 1100000 & 0\% \\ \hline
Day 3 & 1160000 & 1100000 & -5\% \\ \hline
Day 4 & 1200000 & 1140000 & -5\% \\ \hline
Day 5 & 1245000 & 1220000 & -2\% \\ \hline
Day 6 & 1235000 & 1225000 & -1\% \\ \hline
Day 7 & 1275000 & 1175000 & -8\% \\ \hline

\end{tabular}

\end{center}

\section{Conclusion}

In this paper new heuristic was developed for the Site-Dependent Truck and Trailer Routing Problem with Time Windows and Split Deliveries. The heuristic uses a greedy approach for the initial solution construction and then employs elements of local search and tabu search to improve the initial solution. The obtained results are promising as they show improvement in most cases. 

The following work should be directed to the improvement of the speed of the algorithm and to guarantee the best possible results as well. One of the way to improve the algorithm is to use new neighborhood - swap neighborhood, where two customers from different routes can be swapped. Also, there are more constraints that can be relaxed, such as time windows and split deliveries. 

\section{Acknowledgments}
The author is supported by LATNA Laboratory, NRU HSE, RF government grant, ag. 11.G34.31.0057.

\section{References}

\begin{enumerate}
\item Batsyn, M., \& Ponomarenko, A. (2014). \textit{Heuristic for a Real-life Truck and Trailer Routing Problem.} Procedia Computer Science, 31, 778-792. \\doi:10.1016/j.procs.2014.05.328 \\

\item Batsyn, M., \& Ponomarenko, A. (2015). \textit{Heuristic for Site-Dependent Truck and Trailer Routing Problem with Soft and Hard Time Windows and Split Deliveries.} Lecture Notes in Computer Science Machine Learning, Optimization, and Big Data, 65-79. doi:10.1007/978-3-319-27926-8\_7 \\
\end{enumerate}

\end{document}